\documentclass[runningheads]{llncs}
\usepackage{graphicx}
\usepackage{comment}
\usepackage{tikz}
\usepackage{physics}
\usepackage{multirow}
\usepackage{ragged2e}
\usepackage{booktabs, tabularx, wrapfig}
\usepackage{comment}
\usepackage{caption}
\usepackage{float}
\usepackage{url}
\usepackage[breaklinks=true]{hyperref}
\usepackage{breakcites}

\begin{document}
\title{Deep Generative Views to Mitigate Gender Classification Bias Across Gender-Race Groups\thanks{Supported by National Science Foundation (NSF)}}
%
%
\author{Sreeraj Ramachandran\and
Ajita Rattani\thanks{Corresponding author}}
\authorrunning{S. Ramachandran \and A. Rattani}
\titlerunning{Deep Generative Views to Mitigate Gender
Classification Bias}
%
\institute{School of Computing\\
Wichita State University, USA\\
\email{sxramachandran2@shockers.wichita.edu}, 
\email{ajita.rattani@wichita.edu}}
\maketitle              
\begin{abstract}
Published studies have suggested the bias of automated face-based gender classification algorithms across gender-race groups. Specifically, unequal accuracy rates were obtained for women and dark-skinned people. To mitigate the bias of gender classifiers, the vision community has developed several strategies. However, the efficacy of these mitigation strategies is demonstrated for a limited number of races mostly, Caucasian and African-American. Further, these strategies often offer a trade-off between bias and classification accuracy. To further advance the state-of-the-art, we leverage the power of generative views, structured learning, and evidential learning towards mitigating gender classification bias. We demonstrate the superiority of our bias mitigation strategy in improving classification accuracy and reducing bias across gender-racial groups through extensive experimental validation, resulting in state-of-the-art performance in intra- and cross dataset evaluations.

\end{abstract}

\section{Introduction}

Gender is one of the important demographic attributes. Automated gender classification\footnote{Studies in~\cite{bowker,keyes} have shown the inherent problems with gender and race classification. While the datasets used in this study use an almost balanced dataset for the training, it still lacks representation of entire large demographic groups, effectively erasing such categories. We use the gender and race categories defined in these datasets to make comparisons to prior work, not to reinforce or endorse the use of such reductive categories.} refers to automatically assigning gender labels to biometric samples. It has drawn significant interest in numerous applications such as demographic research, surveillance, human-computer interaction, anonymous customized advertisement system, and image retrieval system~\cite{nist_gender,wayman1997large,article1,DBLP:conf/fgr/JainH04}. Companies such as Amazon, Microsoft~\cite{gender_shades}, and many others have released commercial software containing an automated gender classification system from biometric facial images. 

Over the last few years, published research has questioned the fairness of these face-based automated gender classification systems across gender and race~\cite{gender_shades,commercial_bias,bias_color,bias_fairface}.  A classifier is said to satisfy group fairness if subjects in both the protected and unprotected groups have an equal chance of being assigned to the positively predicted class~\cite{10.1145/3194770.3194776}. Existing gender classification systems produce unequal rates for women and people with dark skin, such as African-Americans. Studies in~\cite{BiasDeep,Siddiqui_2022_CVPR} have also evaluated the gender bias of facial analysis based deepfake detection and BMI prediction models. Further, studies in~\cite{10.1007/978-3-030-68793-9_16,9717383} have evaluated the bias of ocular-based attribute analysis models across gender and age-groups.

To mitigate the bias of the gender classification system, several solutions have been developed by the vision community. These solutions are based on regularization strategies~\cite{readme}, attention mechanism~\cite{att_aware_debias} and adversarial debiasing~~\cite{tradeoff1,tradeoff2}, data augmentation techniques~\cite{fair_mixup}, subgroup-specific classifiers~\cite{multitask}, and over-sampling the minority classes using Generative Adversarial Networks(GANs)~\cite{gan_debias}. 

\begin{figure}[!ht]
    \centering
    \includegraphics[scale=1.7]{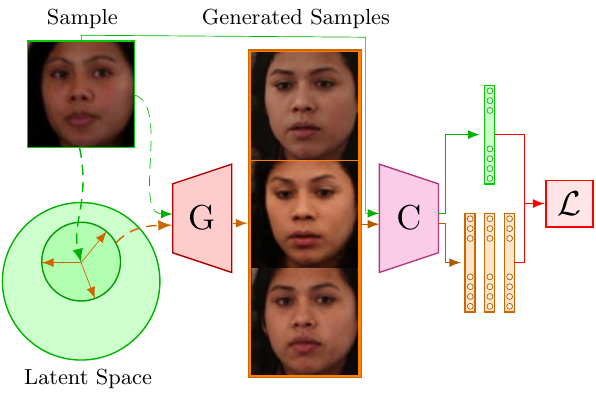}
    \caption{Input samples are projected into a latent space and augmentations are generated. The loss function minimizes the distance between the embedding along with the classification loss.}
    \label{fig:teaser}
    \vspace{-1.5em}
\end{figure}

Often the aforementioned mitigation strategies offer a trade-off~\cite{tradeoff1,tradeoff2} between the fairness and classification task. Further, the efficacy of these strategies are demonstrated on limited number of races; mostly African-American and Caucasian~\cite{gender_shades,bias_color}. A bias mitigation strategy that offer fairness across several inter-sectional subgroups without compromising the gender classification accuracy is still an \textbf{open challenge}~\cite{chlng}.

To further advance the state-of-the-art, this paper proposes a solution that combines the power of structured learning, deep generative views, and evidential learning theory for classifier's uncertainty quantification, for bias mitigation of the gender classification algorithms. We observe that the locally smooth property of the latent space of a Generative Adversarial Network (GAN) model facilitates the generation of the perceptually similar samples. The augmentation strategies that exploit the local geometry of the data-manifold are more powerful in general when used in a consistency regularization style setting. We chose the latent space of the GAN model to be a surrogate space for the data-manifold and therefore sampled augmentations from the latent space called deep generative views. 

Specifically, the facial images of the training set are inverted to the GAN latent space to generate a latent code by training an encoder of the trained StyleGAN. The generated latent code is perturbed to produce variations called \emph{neighboring views} (deep generative views) of the training images. The original training samples along with the neighboring views are used for the gender classifier's training. The regularization term is added to the loss function that enforces the model to learn similar embedding between the original images and the neighboring views. Lastly, a reject option based on uncertainty quantification using evidential deep learning is introduced. The reject option is used to discard samples during the test time based on the quantification of the uncertainty of the classifier's prediction. Figure~\ref{fig:teaser} shows the schema of the proposed approach based on obtaining generative views by projecting the original samples into the latent space.

The proposed bias mitigation strategy has the \textbf{dual advantage} of enhancing the classifier's representational ability as well as the fairness, as demonstrated through extensive experimental evaluations. Also, its generality allows it to be applied to any classification problem, not just face-based gender classification.

In summary, the main \textbf{contributions} of this paper are as follows:

\begin{itemize}
\item A bias mitigation strategy for deep learning-based gender classifiers that leverages and combines the power of GAN to produce deep generative views, structured learning, and evidential learning theory for uncertainty quantification. 

\item We demonstrate the merit of our proposed bias mitigation strategy through experimental analysis on state-of-the-art facial attribute datasets, namely, FairFace~\cite{fairface}, UTKFace~\cite{utkface}, DiveFace~\cite{diveface} and Morph~\cite{morph}, and ocular datasets namely, VISOB~\cite{visob} and UFPR~\cite{ufpr}.

\item Extensive experiments demonstrate the dual advantage of our approach in improving the representational capability as well as the fairness of the gender classification task, thus obtaining state-of-the-art performance over existing baselines most of the time.

\end{itemize}

\begin{figure*}[!hbtp]
    \centering
    \scalebox{0.7}{
\begin{tikzpicture}
\tikzstyle{data_space} = [draw=cyan!70!black,circle, fill=cyan!20!white]
\tikzstyle{latent_space} = [draw=green!70!black,circle, fill=green!20!white, inner sep=15]
\tikzstyle{rep_space} = [draw=magenta!70!black,circle, fill=magenta!20!white, inner sep=15]
\tikzstyle{gen} = [draw=red!70!black,circle, fill=red!20!white]
\tikzstyle{desc} = [draw=magenta!70!black,circle, fill=magenta!20!white]
\tikzstyle{enc} = [draw=magenta!70!black,circle, fill=magenta!20!white]
\tikzstyle{cls} = [draw=magenta!70!black,circle, fill=magenta!20!white]

\node [align=left] at (0.1,-0.4) {\scriptsize Given: $X$ \\ \scriptsize Learn: $G$
};
\node [latent_space] (v2) at (0.6,-2.1) {};
\node at (0.5,-1.7) {$\scriptstyle z\vdot$};

\node at (0.5,-3.1) {\scriptsize Latent space};
\draw [gen,path picture={
      \node at (path picture bounding box.center) {
        {G}};}] (1.9,-1.8) node (v1) {} -- (1.9,-2.4) -- (2.6,-2.7) -- (2.6,-1.5) -- cycle;

\draw [gen,path picture={
      \node at (path picture bounding box.center) {
        {D}};}] (5.4,-1.8) node (v1) {} -- (5.4,-2.4) -- (4.7,-2.7) -- (4.7,-1.5) -- cycle;
\draw [-latex] (0.5634,-1.6918) .. controls (1,-2) and (1.5,-2.1) .. (1.9,-2.1);
\draw [data_space]  plot[smooth cycle, tension=.7] coordinates {(3,-2.2) (3.1,-2.7) (3.6,-2.5) (4.1,-2.7) (4.4,-2.3) (3.9,-2) (3.8,-1.5) (3.3,-1.7) (3.3,-2.1)};
\node at (3.5,-2.3) {$\scriptstyle \vdot  x'$};
\node at (3.6,-1.7) {$\scriptstyle \vdot x$};
\node at (3.6,-3) {\scriptsize Data space};
\node at (2.3,-1.3) {\scriptsize Generator};
\node at (5.2,-1.3) {\scriptsize Discriminator};
\draw [-latex] (2.6,-2.1) .. controls (2.9,-2.3) and (3.2,-2.1) .. (3.3753,-2.3399);
\draw [-latex] (3.3694,-2.3373) .. controls (3.7,-2) and (4.2,-2) .. (4.7,-2.4);
\draw [-latex] (3.5212,-1.6938) node (v5) {} .. controls (3.8,-1.3) and (4.3,-1.5) .. (4.7,-1.9);
\node [draw,rectangle] (v3) at (6.5,-2.1) {$\mathcal{L}_{GAN}$};
\draw [-latex] (5.4,-2.1) -- (v3);
\node (v4) at (3.5,-0.9) {\scriptsize $x \in X$};
\draw [-latex] (v4) -- (v5.center);

\node [align=left] at (9.2277,-0.1171) {\scriptsize Given: $G, X$ \\
\scriptsize Learn: $E$
};
\node (v6) at (8.2,-2.1) {$x \in X$};

\draw [gen,path picture={
      \node at (path picture bounding box.center) {
        {E}};}] (9.965,-1.805) node (v1) {} -- (9.965,-2.405) -- (9.265,-2.705) -- (9.265,-1.505) -- cycle;
\draw [-latex] (v6) -- (9.265,-2.105);
\node[latent_space] at (11.065,-2.105) {};
\node at (11.065,-2.405) {$\scriptstyle \vdot z$};
\draw [-latex] (9.965,-2.105) .. controls (10.415,-2.005) and (10.765,-2.105) .. (10.979,-2.4001);
\node at (9.565,-1.305) {\scriptsize Encoder};
\node at (11.015,-3.005) {\scriptsize Latent space};

\draw [gen,path picture={
      \node at (path picture bounding box.center) {
        {G}};}] (12.265,-1.805) node (v1) {} -- (12.265,-2.405) -- (12.965,-2.705) -- (12.965,-1.505) -- cycle;

\draw [-latex] (10.9912,-2.3957) .. controls (11.265,-2.105) and (11.665,-2.055) .. (12.269,-2.1239);
\node at (12.815,-1.355) {\scriptsize Generator};
\draw [data_space]  plot[smooth cycle, tension=.7] coordinates {(13.47,-2.205) (13.57,-2.705) (14.07,-2.505) (14.57,-2.705) (14.87,-2.305) (14.37,-2.005) (14.27,-1.505) (13.77,-1.705) (13.77,-2.105)};

\node at (14.115,-1.905) {$\scriptstyle \vdot x'$};
\draw [-latex] (12.965,-2.105) .. controls (13.215,-1.805) and (13.465,-1.755) .. (13.9777,-1.9471);
\node at (14.115,-3.005) {\scriptsize Data space};
\node [draw, rectangle] (v8) at (15.715,-2.105) {$\mathcal{L}_{e4e}$};
\draw [-latex] (13.9768,-1.9461) .. controls (14.0158,-1.5359) and (14.7391,-1.9778) .. (15.2596,-2.1347);

\node [align=left] at (0.8,-6.9) {\scriptsize Given: $G, E, X$ \\
\scriptsize Generate: $\mathcal{N}(x), x \in X$
};
\draw [data_space]  plot[smooth cycle, tension=.7] coordinates {(-0.6,-5) (-0.5,-5.5) (0,-5.3) (0.5,-5.5) (0.8,-5.1) (0.3,-4.8) (0.2,-4.3) (-0.3,-4.5) (-0.3,-4.9)};
\node at (-0.1,-4.6) {$\scriptstyle  x\vdot$};

\draw [gen,path picture={
      \node at (path picture bounding box.center) {
        {E}};}] (2,-4.6) node (v1) {} -- (2,-5.2) -- (1.3,-5.5) -- (1.3,-4.3) -- cycle;
\node [latent_space] (v2) at (3.2,-4.9) {};
\draw [gen,path picture={
      \node at (path picture bounding box.center) {
        {G}};}] (4.4,-4.6) node (v1) {} -- (4.4,-5.2) -- (5.1,-5.5) -- (5.1,-4.3) -- cycle;

\draw [data_space]  plot[smooth cycle, tension=.7] coordinates {(5.6,-4.9) (5.7,-5.4) (6.2,-5.2) (6.7,-5.4) (7,-5) (6.5,-4.7) (6.4,-4.2) (5.9,-4.4) (5.9,-4.8)};
\node at (3,-5.3) {$\scriptstyle \vdot z$};
\node at (3.3,-5) {$\scriptstyle \vdot z'$};
\draw [-stealth, densely dotted, thick, draw=orange] (2.9419,-5.2854) .. controls (2.9205,-5.1095) and (3.032,-5.0108) .. (3.1951,-5.028);
\node at (2.8518,-5.0195) {$\scriptscriptstyle T_z$};

\draw [-latex] (0.0019,-4.5875) .. controls (0.4633,-4.4135) and (0.7658,-4.9354) .. (1.3029,-4.9657);
\draw [-latex] (2.014,-4.9127) .. controls (2.3241,-4.8598) and (2.4678,-5.3288) .. (2.9104,-5.3006);
\draw [-latex] (3.1997,-5.0267) .. controls (3.3858,-4.634) and (3.9179,-4.9234) .. (4.4139,-4.9182);
\node at (6.124,-5.0164) {$\scriptstyle \vdot\mathcal{N}(x)$};

\draw [-latex] (5.1062,-4.9182) .. controls (5.4109,-4.9699) and (5.4988,-4.7219) .. (5.7934,-5.0215);
\node at (0.0732,-5.9148) {\scriptsize Data space};
\node at (3.1793,-5.9935) {\scriptsize Latent space};
\node at (6.2503,-5.9585) {\scriptsize Data space};
\node at (1.5868,-4.0074) {\scriptsize Encoder};
\node at (4.6754,-4.0074) {\scriptsize Generator};


\node [align=left] at (9.5389,-7.0384) {\scriptsize Given: $X, \mathcal{N}(X)$ \\
\scriptsize Learn: $F$
};

\draw [data_space]  plot[smooth cycle, tension=.7] coordinates {(8.4823,-5.0714) (8.5823,-5.5714) (9.0823,-5.3714) (9.5823,-5.5714) (9.8823,-5.1714) (9.3823,-4.8714) (9.2823,-4.3714) (8.7823,-4.5714) (8.7823,-4.9714)};

\draw [gen,path picture={
      \node at (path picture bounding box.center) {
        {F}};}] (10.8823,-4.7714) node (v1) {} -- (10.8823,-5.3714) -- (10.1823,-5.6714) -- (10.1823,-4.4714) -- cycle;

\node at (8.9834,-4.7815) {$\scriptstyle \vdot x$};
\node at (9.352,-5.2059) {$\scriptstyle \vdot \mathcal{N}(x)$};
\node [rep_space] at (12.1526,-5.0281) {};
\node at (12.0727,-5.4529) {$\scriptstyle \vdot F(x)$};
\node at (12.2491,-4.823) {$\scriptstyle \vdot F(\mathcal{N}(x))$};
\node [draw, rectangle] at (13.5898,-4.9714) {$\mathcal{L}_{cls}$};
\node at (14.1753,-4.9714) {$+$};
\node [draw, rectangle] at (14.7684,-4.9714) {$\mathcal{L}_{\mathcal{N}}$};
\draw [-latex] (8.9012,-4.7717) .. controls (9.2786,-4.5478) and (9.6435,-4.5403) .. (10.1758,-5.1106);
\draw [-latex] (9.0201,-5.1942) .. controls (9.1721,-4.9737) and (9.6512,-4.9661) .. (10.1834,-5.1258);
\draw [-latex] (10.8677,-5.0725) .. controls (11.0502,-4.8064) and (11.324,-4.6696) .. (11.6965,-4.8216);
\draw [-latex] (10.8677,-5.0725) .. controls (11.1339,-5.3007) and (11.4684,-5.4679) .. (11.765,-5.4451);
\draw [-latex] (11.765,-5.4451) .. controls (12.1452,-5.0118) and (12.6393,-4.9736) .. (13.1488,-5.0193);
\draw [-latex, draw=orange] (11.6889,-4.814) .. controls (12.1527,-3.9776) and (13.9776,-3.9472) .. (14.7,-4.7);
\draw [-latex, draw=orange] (11.7574,-5.4451) .. controls (12.2972,-5.9925) and (14.3654,-6.1599) .. (14.7684,-5.2246);
\node at (9.135,-6.1838) {\scriptsize Data space};
\node [align=center] at (12.1378,-6.2819) {\scriptsize Representation \\ \scriptsize space};
\node at (10.3126,-4.0838) {\scriptsize Classifier};

\draw [dashed, draw=gray] (-1,-3.4) -- (16,-3.4);
\draw [dashed, draw=gray] (7.4,-0.1) -- (7.4,-6.9);
\node at (-0.9,-0.3) {(A)};
\node at (8,0) {(B)};
\node at (-1,-4) {(C)};
\node at (8,-4) {(D)};

\end{tikzpicture}}
    \caption{Overview of the proposed method. (A) Training a GAN to learn the input image data distribution, $X$. (B) Training an Encoder model, $E$ to learn to project input image, $x$ to the latent space. (C) Generating Neighbour Views by first projecting input image into latent space using $E$, applying perturbation in latent space, then using $G$ to generate. (D) Training the Classifier, $F$ using both the input image and its neighbors.} 
    \label{fig:nsl_teaser}
\end{figure*}
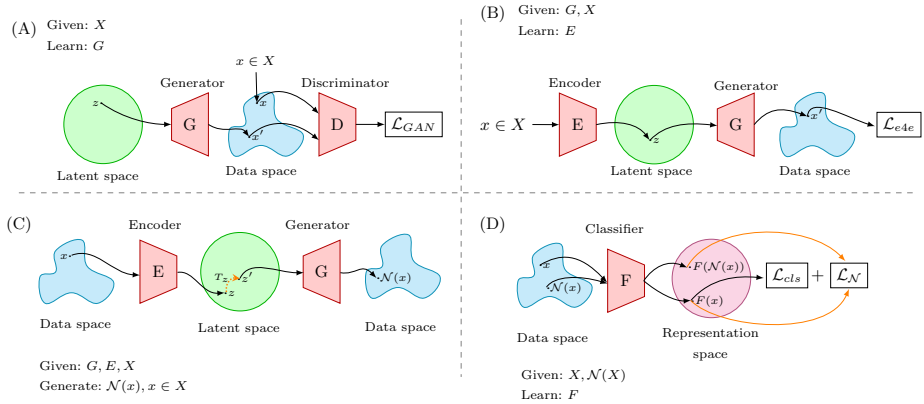

\section{Related Work}

\noindent\textbf{Fairness in Gender Classification}:
In~\cite{gender_shades}, authors evaluated the fairness using commercial SDKs and observed least accuracy rates for dark-skinned females. Studies in~\cite{gender_age_transfer,Ryu18} proposed data augmentation and two-fold transfer learning for measuring and mitigation bias in deep representation. Attribute aware filter drop was proposed in~\cite{att_aware_filter} and regularization strategy in~\cite{tartaglione} are used to unlearn the dependency of the model on sensitive attributes. 
In ~\cite{multitask}, a multi-task Convolution Neural Network (MTCNN) is proposed, that use a joint dynamic loss to jointly classify race, gender and age to minimize bias. A data augmentation strategy called fair mixup is proposed in~\cite{fair_mixup}, that regularize the model on interpolated distributions between different sub-groups of a demographic group. An auto-encoder that disentangle data representation into latent sub-spaces is proposed by~\cite{readme}. 
Using GANs to generate balanced training set images w.r.t protected attributes was proposed in~\cite{gan_debias}. Recently OpenAI released its large scale language-image pretraining model called CLIP~\cite{clip} which was trained contrastively on 400M image-text pairs and demonstrated its exceptional performance as well as bias on both zero-shot and fine-tuning setting on gender classification tasks.\\

\noindent \textbf{GANs for real image editing}: 
Advancements in GANs~\cite{gan,gan1,progan,bigan,stylegan,stylegan2,stylegan2-ada} enable us to produce increasingly realistic images that can replicate variations in training data. Specifically, the learnt intermediate latent space of the StyleGAN~\cite{stylegan,stylegan2,stylegan2-ada} represents the distribution of the training data more efficiently than the traditional Gaussian latent space. By leveraging the disentangling properties of the latent space as shown in~\cite{editing4,walking1,editing2,editing1,e4e,psp}, extensive image manipulations could be performed. Once we invert the input image into the learned latent space, we can modify real images in a similar way. High-quality inversion methods (i) can properly reconstruct the given image, and (ii) produce realistic and semantically meaningful edits. Optimization methods~\cite{invert3,invert2,invert1} optimize the latent code by minimizing error for an individual image, whereas faster encoder-based methods~\cite{invert4,invert5,e4e} train an encoder to map images to the latent space. 

Once inverted, manipulations can be done on the given image by finding “walking” directions that control the attribute of interest. Studies in ~\cite{walking1,ganalyze,walking_bias} use supervised techniques to find these latent directions specific to attributes. Whereas studies in~\cite{editing2,walking_un1,walking_un2} find directions in an unsupervised manner, requiring manual identification of the determined directions later and~\cite{sefa} using a closed-form factorization algorithm for identifying top semantic latent directions by directly decomposing the pre-trained weights. \\

\noindent \textbf{Structured Learning, Ensembling of Deep Generative Views}: 
Studies in~\cite{neural_graph_learning,ssl_gcn,goodfellow_adverarial,miyato} have shown that combining feature inputs with a structured signal improves the overall performance of a model in various scenarios such as Graph Learning and Adversarial learning. These structured signals either implicit (adversarial) or explicit (graph) are used to regularize the training while the model learns to accurately predict (by minimizing supervised loss)~\cite{nsl} by maintaining the input structural similarity through minimizing a neighbor loss. Recently ~\cite{ensembling_views} proposed a test-time ensembling with GAN-based augmentations to improve classification results. 


\section{Proposed Approach}
Figure~\ref{fig:nsl_teaser} illustrates the overview of the proposed approach. The process involves training a generative model to produce different views of a given input image. The training images are projected to the latent space of the generator. The variations to the input images are produced using augmentation in the latent space. These generative views act as a structured signal and are used along with the original images to train the downstream gender classifier. Therefore, by injecting prior knowledge into the learning process through the learned latent space and by enforcing local invariance properties of the manifold when used as a consistency regularizer, the classifier's performance is significantly improved. This helps propagate the information contained in the labeled samples to the unlabeled neighboring samples. The proposed strategy of leveraging neighbor views is used during classifier's training, the test images are used as it is during the test-time. 

\subsection{GAN Preliminaries}
A GAN~\cite{gan} consists of a generator network~($G$) and a discriminator network~($D$) involved in a min-max game, where $G$ tries to fool $D$ by producing realistic images from a latent vector $z$ and D gets better at distinguishing between real and generated data. We employ the StyleGAN2~\cite{stylegan2} generator in this study. Previous works have shown that the intermediate $W$ space, designed to control the "style" of the image is better able to represent images than the original latent code $z$ with fine-grained control and better disentanglement. In a StyleGAN2 generator, a mapping network $M$ maps $z$ to $w \in W$ and $G$ generates the image $x$, given $w$. i.e $x = G(M(z))$.

\subsection{Projecting Images into GAN Latent Space}
To alter an image $x$ with the generator, we must first determine the appropriate latent code that generates $x$. GAN inversion refers to the process of generating the latent code for a given image $x$. For GAN inversion, there are optimization-based methods as well as encoder-based methods that offer various trade-offs between edit-ability, reconstruction distortion, and speed. We specifically use the \emph{encoder4editing} method (e4e)\cite{e4e}. e4e use adversarial training to map a given real image to a style code composed of a series of low variance style vectors, each close to the distribution of $W$. We can generate the reconstruction, $x'$, which is closer to the original $x$ using the given style vectors and the generator, G. 

\subsection{Generating Views using Pretrained GAN}
For generating views, we use existing latent editing methods. We specifically employ SeFA~\cite{sefa}, a closed-form factorization technique for latent semantic discovery that decomposes the pre-trained weights directly. Unsupervised, SeFa determines the top $k$ semantic latent directions. To produce alternate views, we randomly select an arbitrary set of semantic latent directions and sample distances from a Gaussian distribution for latent space traversal for each image $x$. The generator then uses these updated style vectors to generate the views. In the case of latent vectors occurring in poorly defined i.e., warped region of latent space which may produce non-face images, a simple MTCNN-based~\cite{mtcnn} face detector is utilized to screen out the non-faces.

\subsection{Structured Learning on Deep Generative Views}
Let $x_i \in X$ be the input sample. We obtain its corresponding latent style vector $w_i$ using e4e. We may apply SeFA on $w_i$ to obtain $N(w_i)=\{w^1_i,\ldots,w^m_i\}$, where $m$ is a hyperparameter representing the number of neighbours, the neighbouring latent vectors. Using the pretrained StyleGAN generator we may produce the neighbouring views $N(x_i)=G(N(w_i))$.

During training each batch will contain samples with pairs of original sample $x_i$ and the generated neighbours $N(x_i)$. Both $x_i$ and $N(x_i)$ are used in the forward pass but only $x_i$ is backpropagated and used for calculating batch statistics in the normalization layers. This is because of the distributional difference between the real and generated images. The total loss, $\mathcal{L}_{total}$ is given by
\begin{equation}
    \mathcal{L}_{total} = \mathcal{L}_{cls}(y_i,y'_i) + \alpha\displaystyle\sum_{x_j \in \mathcal{N}(x_i)} \mathcal{L}_{\mathcal{N}}(h_{\theta}(x_i),h_{\theta}(x_j))
\end{equation}
where $\mathcal{L}_{cls}(y_i,y'_i)$ is the classification loss, $\alpha$ is a hyperparameter, $\mathcal{L}_{\mathcal{N}}$ is any distance function that can be used to calculate the distance between the sample embedding and the neighbour embedding and $h_{\theta}(x)$ is the sample embedding from the neural network. For our experiments, we use Jensen-Shannon Divergence~\cite{jsdiv} to calculate the distance between the sample embedding and the neighbour embedding.

\subsection{Evidential Deep Learning for Quantifying Uncertainty}
In standard classifier training where prediction loss is minimized, the resultant model is ignorant of the confidence of its prediction. A study in~\cite{edl} proposed explicit modeling of uncertainty using subjective logic theory by applying a Dirichlet distribution to class probabilities. Thus, treating model predictions as subjective opinions, and learning the function that collects the evidence leading to these opinions from data using a deterministic neural net. The resulting predictor for a multi-class classification problem is another Dirichlet distribution, the parameters of which are determined by a neural net's continuous output. 

Let us assume that $\alpha_i = \alpha_{i1},\ldots,\alpha_{ik}$ is the parameters of a Dirichlet distribution for the classification of a sample $i$, then $(\alpha_{ij} - 1)$ is the total evidence, $e_j$ estimated by the network for the assignment of the sample $i$ to the $j^{th}$ class. The epistemic uncertainty can be calculated from the evidences $e_i$ using 
\begin{equation}\label{eq:unc}
    u = \frac{K}{S_i}
\end{equation}
where $K$ is the number of classes and $S_i = \sum_{j=1}^{K}(e_j + 1)$
\cite{edl} proposes the following loss function in this scenario: 
\begin{equation}\label{eq:loss_i}
    \mathcal{L}_i = \sum_{j=1}^{K}(y_{ij}-\hat{p}_{ij})^2 + \frac{\hat{p}_{ij}(1-\hat{p}_{ij})}{S_i+1}
\end{equation}
\begin{equation}\label{eq:loss}
    \mathcal{L}_{edl} = \sum_{i=1}^{N}\mathcal{L}_{i} + \lambda_{t}\sum_{i=1}^{N}KL[D(\boldsymbol{p_i|\tilde{\alpha}_i})||D(\boldsymbol{p}_i|\langle1,\dots,1\rangle)]
\end{equation}
where $\boldsymbol{y}_i$ is the one-hot vector encoding of ground-truth and $\boldsymbol{p}_i$ is the class-assignment probabilities given by $\hat{p}_k=\alpha_k/S$.

We use the estimated uncertainty as the basis for rejection, similar to reject-option-based classification~\cite{reject_option}, where a test sample in low confidence region is rejected based on the model's confidence values. The provided model of uncertainty is more detailed than the point estimate of the standard softmax-output networks and can handle out-of-distribution queries, adversarial samples as well as samples belonging to the critical region. Therefore a high uncertainty value of a test sample suggests the model's under-confidence. And we use this uncertainty estimate as the basis for rejecting samples during test time rather than using the softmax probabilities.

\section{Experiments and Results}
\subsection{Datasets used}
For all our experiments we used the FairFace~\cite{fairface} as our training dataset. Testing was done on the test set of the FairFace as well as DiveFace~\cite{diveface}, UTKFace~\cite{utkface} and Morph~\cite{morph} datasets.

Table~\ref{tab:datasets} shows the characteristics of these datasets used in our study.

\begin{table}[!ht]
    \centering
    \caption{Datasets}
    \label{tab:datasets}
    \begin{tabular}{llp{8cm}}
    \toprule
    \multicolumn{1}{c}{\textbf{Dataset}} & \multicolumn{1}{c}{\textbf{Images}} & \multicolumn{1}{c}{\textbf{Races}} \\ \midrule
    FairFace & $100k$ & White, Black, Indian, East Asian, Southeast Asian,~Middle Eastern, Latino Hispanic \\
    DiveFace & $150k$ & East Asian, Sub-Saharan,South Indian, Caucasian \\
    UTKFace & $20k$ & White, Black, Indian, Asian \\
    Morph & $55k$ & White, Black\\
    \bottomrule
    \end{tabular}
\end{table}

\subsection{Gender Classification}
For our experiments on gender classification, we used FairFace as our training dataset. The FairFace dataset was also used for training the StyleGAN2 generator as well as the e4e encoder that was subsequently used to generate the neighbouring views.


\subsubsection{\textbf{Generating Neighbouring Views}}
\begin{figure*}
    \centering
    \includegraphics[width=0.98\textwidth]{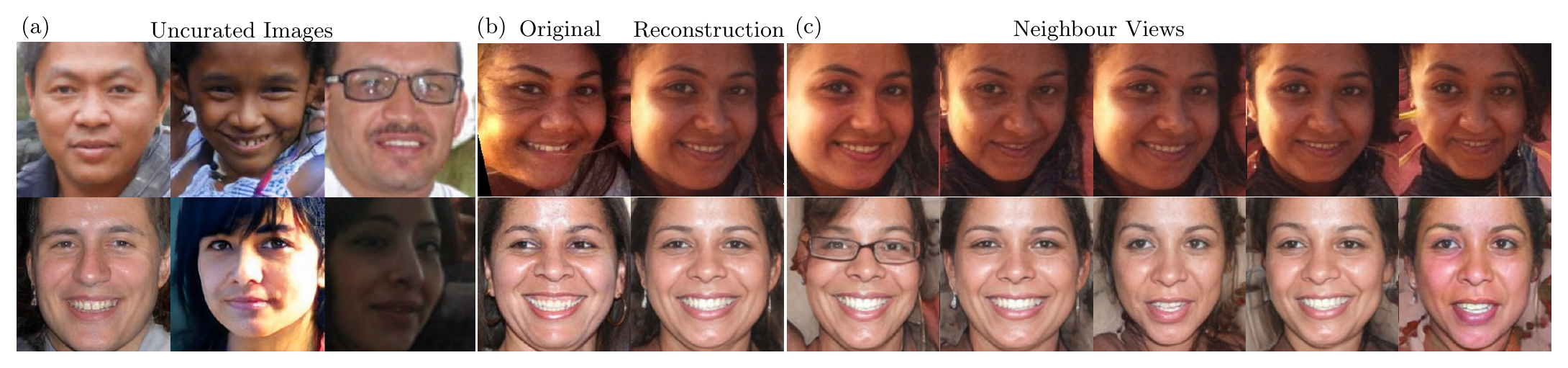}
    \caption{(a) Uncurated images generated by StyleGAN2 trained on FairFace. (b) Images reconstructed using the trained e4e encoder and pretrained StyleGAN2 generator. (c) Generative views created by selecting top semantic latent directions with SeFA and randomly sampling along those directions.}
    \label{fig:imgs_recons_vars}
    \vspace{-0.2em}
\end{figure*}

: For pretrained StyleGAN2, we use the official StyleGAN2-ADA implementation pretrained on high quality FFHQ-256 face dataset~\cite{stylegan} and used transfer learning on the FairFace training set with images resized to $256 \times 256$. We obtained a Fréchet inception distance (FID)~\cite{fid} of $4.29$, similar perceptual quality as of FFHQ on StyleGAN~\cite{stylegan}. The uncurated images generated by the trained generator are shown in Figure~\ref{fig:imgs_recons_vars}.

For pretraining e4e, we use the official e4e implementation with the pretrained StyleGAN2 generator from above and the FairFace training set. Figure \ref{fig:imgs_recons_vars} shows uncurated reconstructions by the encoder. Finally, for generating the views, the training set images are first inverted using e4e and then we use SeFA to choose the top $k$ semantic latent directions. We pick distances in both positive and negative directions randomly. For our experiments we initially generate 56 neighbors per image, to have a wide variation of views. We then use a pretrained MTCNN face detector to detect faces on the generated images and remove non-faces to ensure a clean dataset sampled from the well-defined region of latent space. Figure~\ref{fig:imgs_recons_vars} shows examples of generated variations. 

\subsubsection{\textbf{Training the classifier}}: For our experiments, we use a baseline \emph{EfficientNetV2-L} pretrained on ImageNet as a gender classification model. Our proposed method based on deep generative views and structured learning, denoted by \emph{Neighbour Learning~(NL)}, is applied to this model. We also compare our results with a sensitive-attribute aware multi-task classifier~(MT)~\cite{multitask}, with race as the secondary attribute. 

Further, we combine both NL and MT and evaluate their performance (third configuration). And finally, our last configuration which also has the ability to predict uncertainty, combines NL, and MT and replaces the final classification head of the gender classifier with an evidential layer (EDL) and replaced cross-entropy loss with an evidential loss. 

For all our experiments, we use an RMSProp optimizer with a cosine annealing schedule with an initial learning rate of $4\times 10^{-4}$ and weight decay of $1\times 10^{-5}$. We use a batch size of $128$ across 2 RTX8000 GPUs, with label smoothing of $0.1$, and \emph{autoaugment} policy for data augmentation. For NL, $\alpha$ is set to $2$, and the number of neighbors, $m$ to $7$ for the final configuration. We also apply \emph{lazy regularization} to speed up the training, inspired by StyleGAN, where we apply the costly NL regularization every $n$ batch. For our final model, we set this hyperparameter value to $2$. For evaluation, we used the \emph{Degree of Bias~(DoB)}~\cite{dob}, the standard deviation of classification accuracy across different subgroups, as well as \emph{Selection Rate (SeR)}~\cite{ser}, the ratio of worst to best accuracy rate, as a metric for evaluation of the fairness.

Table \ref{tab:results} show the results on the FairFace validation set. The empirical results suggest that the proposed method NL improves the fairness of the model as well as the overall accuracy, outperforming both baseline and multi-task aware setup~(MT). Combining NL with MT and evidential deep learning~(EDL) further improves the performance by reducing the DoB to $1.62$ and with an accuracy of $94.70\%$. To compare with the state of the art as well as to evaluate the generality of the method, we applied the same to the vision tower of the CLIP~\cite{clip} model. We used the ViT-L/14 version of the model with pretrained weights and added a final linear classification head for the configuration. Applying our method to the CLIP model~\cite{clip}, an already SOTA method, improved the DoB while maintaining the accuracy. As most studies constraint the bias evaluation to a simple white vs non-white subjects, we restrict our comparisons to SOTA methods~\cite{lp_insta,clip} that evaluate fairness across multiple-race groups on FairFace and observed our method either outperformed or enhanced the models that were pretrained on huge datasets such as WebImageText($400M$)~\cite{clip} and Instagram ($3.5B$)~\cite{lp_insta}.

\begin{table*}[hbtp]
    \centering
    \caption{Gender classification results of the proposed method across gender-race groups when trained and tested on FairFace.} 
    \label{tab:results}
    \resizebox{\textwidth}{!}{
    \begin{tabular}{lccccccccccc} 
    \toprule
    \multicolumn{1}{c}{\textbf{Configuration}} & \textbf{Gender} & \textbf{Black} & \begin{tabular}[c]{@{}c@{}}\textbf{East}\\\textbf{Asian}\end{tabular} & \textbf{Indian} & \begin{tabular}[c]{@{}c@{}}\textbf{Latino}\\\textbf{Hispanic}\end{tabular} & \begin{tabular}[c]{@{}c@{}}\textbf{Middle}\\\textbf{Eastern}\end{tabular} & \begin{tabular}[c]{@{}c@{}}\textbf{Southeast}\\\textbf{Asian}\end{tabular} & \textbf{White} & \textbf{Average $\uparrow$} & \textbf{DoB $\downarrow$} & \textbf{SeR $\uparrow$} \\ \midrule
    
    \multirow{2}{*}{A Baseline} & Male & 91.24 & 94.08 & 95.88 & 93.06 & 97.54 & 92.93 & 94.30 & \multirow{2}{*}{94.27} & \multirow{2}{*}{2.01} & \multirow{2}{*}{91.96} \\
     & Female & 89.70 & 94.95 & 94.50 & 96.02 & 95.96 & 96.18 & 94.18 &  &  &  \\\cmidrule[0.01pt](lr){2-12}
     
    \multirow{2}{*}{B   + NL (LR=2)} & Male & 91.24 & 95.24 & 96.15 & 94.83 & 97.79 & 94.01 & 95.63 & \multirow{2}{*}{94.67} & \multirow{2}{*}{1.67} & \multirow{2}{*}{\textbf{93.78}} \\
     & Female & 91.81 & 94.70 & 94.89 & 95.18 & 96.21 & 95.29 & 93.04 &  &  &  \\\cmidrule[0.01pt](lr){2-12}

     \multirow{2}{*}{C   + NL (No LR)} & Male & 91.74 & 95.62 & 96.41 & 95.08 & 97.91 & 94.29 & 95.45 & \multirow{2}{*}{\textbf{95.06}} & \multirow{2}{*}{1.67} & \multirow{2}{*}{93.68} \\
     & Female & 91.68 & 95.08 & 95.94 & 96.63 & 95.96 & 95.59 & 93.87 &  &  &  \\\cmidrule[0.01pt](lr){2-12}
     
    \multirow{2}{*}{D   + MT} & Male & 91.61 & 94.85 & 94.82 & 94.45 & 97.91 & 94.56 & 95.99 & \multirow{2}{*}{94.58} & \multirow{2}{*}{1.73} & \multirow{2}{*}{92.83} \\
     & Female & 90.89 & 94.05 & 95.28 & 95.90 & 96.21 & 93.82 & 93.77 &  &  &  \\\cmidrule[0.01pt](lr){2-12}
     
    \multirow{2}{*}{E   + MT + NL} & Male & 91.86 & 94.98 & 96.81 & 93.69 & 97.79 & 93.20 & 94.92 & \multirow{2}{*}{94.59} & \multirow{2}{*}{1.66} & \multirow{2}{*}{93.48} \\
     & Female & 91.41 & 94.95 & 95.41 & 95.66 & 95.45 & 94.12 & 94.08 &  &  &  \\\cmidrule[0.01pt](lr){2-12}
     
    \multirow{2}{*}{F    + MT + NL + EDL} & Male & 91.86 & 94.85 & 95.35 & 94.58 & 97.79 & 92.65 & 94.83 & \multirow{2}{*}{94.70} & \multirow{2}{*}{\textbf{1.62}} & \multirow{2}{*}{93.62} \\
     & Female & 91.55 & 95.73 & 95.41 & 96.02 & 95.71 & 94.41 & 95.02 &  &  &  \\\midrule
     
    \multirow{2}{*}{G CLIP + Linear} & Male & 94.99 & 96.78 & 97.21 & 96.72 & 98.77 & 96.05 & 97.78 & \multirow{2}{*}{\textbf{96.76}} & \multirow{2}{*}{1.10} & \multirow{2}{*}{95.36} \\
     & Female & 94.19 & 96.63 & 97.25 & 97.83 & 96.71 & 96.62 & 96.47 &  &  &  \\\cmidrule[0.01pt](lr){2-12}
     
    \multirow{2}{*}{H   + NL} & Male & 95.12 & 96.78 & 97.21 & 96.72 & 98.52 & 95.92 & 97.15 & \multirow{2}{*}{96.70} & \multirow{2}{*}{\textbf{0.99}} & \multirow{2}{*}{\textbf{95.87}} \\
     & Female & 94.45 & 96.90 & 96.59 & 97.71 & 97.22 & 96.18 & 96.99 &  &  &  \\
      \cmidrule[0.01pt](lr){2-12}
     
    \multirow{2}{*}{I  + MT + NL + EDL} & Male & 95.37 & 96.78 & 97.08 & 97.10 & 98.89 & 95.92 & 97.15 & \multirow{2}{*}{96.70} & \multirow{2}{*}{1.00} & \multirow{2}{*}{95.38} \\
     & Female & 94.32 & 96.63 & 96.46 & 97.11 & 97.22 & 96.75 & 96.78 &  &  &  \\
     
     \midrule
     \midrule
     \multirow{2}{*}{Instagram + Linear~\cite{lp_insta}} & Male & 92.50 & 93.40 & 96.20 & 93.10 & 96.00 & 92.70 & 94.80 & \multirow{2}{*}{93.77} & \multirow{2}{*}{1.73} & \multirow{2}{*}{93.66} \\
     & Female & 90.10 & 94.30 & 95.00 & 94.80 & 95.00 & 94.10 & 91.40 &  &  &  \\
    \bottomrule
    \end{tabular}
    }
\end{table*}

\textbf{Cross dataset} evaluations were done on UTKFace, DiveFace, and Morph Datasets. Table \ref{tab:cross_dataset} shows cross-dataset results. In the calculation of DoB for datasets where the test set is not balanced across demographic groups, a weighted standard deviation was used instead of a standard deviation.  For DiveFace and Morph, various configurations improved the overall fairness over the baseline whereas it was not the case for UTKFace. It must also be noted that across all datasets the proposed method improved the overall classification accuracy regardless.

\begin{table}
\centering
\caption{Cross-dataset evaluation of the proposed model.}
\label{tab:cross_dataset}
\resizebox{\textwidth}{!}{

\begin{tabular}{lccccccccc} 
\toprule
                     & \multicolumn{3}{c}{\textbf{UTKFace}}                                               & \multicolumn{3}{c}{\textbf{DiveFace}}                                              & \multicolumn{3}{c}{\textbf{Morph}}                                                  \\ 
\cmidrule(l){2-10}
                     & \textbf{DoB $\downarrow$} & \textbf{Avg. Acc $\uparrow$} & \textbf{SeR $\uparrow$} & \textbf{DoB $\downarrow$} & \textbf{Avg. Acc $\uparrow$} & \textbf{SeR $\uparrow$} & \textbf{DoB $\downarrow$} & \textbf{Avg. Acc $\uparrow$} & \textbf{SeR $\uparrow$}  \\ 
\midrule
Baseline             & \textbf{1.96}                      & 94.67                        & \textbf{92.60}                   & 0.74                      & 98.45                        & 97.71                   & 7.67                      & 96.26                        & 74.89                    \\
MT                   & 3.01                      & 94.25                        & 88.89                   & 0.78                      & 98.34                        & 97.67                   & 10.01                     & 95.02                        & 69.97                    \\
NL (LR=2)            & 2.55                      & 94.54                        & 90.11                   & \textbf{0.49}             & 98.49                        & \textbf{98.48}          & 8.99                      & 95.95                        & 70.68                    \\
NL (No LR)           & 2.26                      & \textbf{94.76}                        & 91.55                   & 0.51                      & \textbf{98.60}               & 98.39                   & 7.72                      & \textbf{96.41}               & 74.84                    \\
MT + NL              & 2.31                      & 94.47                        & 90.91                   & 0.77                      & 98.40                        & 97.52                   & 6.69                      & 96.21                        & 78.06                    \\
MT + NL + EDL        & 2.27                      & 94.50                        & 91.53                   & 0.83                      & 98.37                        & 97.43                   & \textbf{6.26}             & 96.32                        & \textbf{78.98}           \\
\midrule
CLIP + LP            & 1.86                      & \textbf{96.58}               & 92.91                   & 0.68                      & \textbf{99.08}                        & 97.90                   & \textbf{0.89}                      & \textbf{99.46}                        & \textbf{97.04}                    \\
CLIP + NL            & \textbf{1.60}             & 96.47                        & \textbf{94.59}          & \textbf{0.62}                      & 99.02                        & \textbf{98.09}                   & 1.39                      & 99.19                        & 95.43                    \\
CLIP + MT + NL + EDL & 2.04                      & 96.52                        & 92.85                   & 0.69                      & 99.04                        & 97.81                   & 1.10                      & 99.26                        & 96.33                    \\
\bottomrule
\end{tabular}
}
\end{table}

\subsubsection{\textbf{Rejecting Samples based on Uncertainty as Threshold}} For evidential learning, we replace the classification head such that, it outputs parameters of a Dirichlet Distribution. We also replace the classification loss $\mathcal{L}_{cls}$ with the evidential loss term defined by Equation~\ref{eq:loss}. For our experiments, removing the annealing coefficient $\lambda_{t}$ worked better. We may calculate the uncertainty of the prediction using Equation~\ref{eq:unc}. This uncertainty is used as a threshold to reject/accept the prediction of the model.
\begin{figure}[!ht]
    \centering
    \includegraphics[scale=0.8]{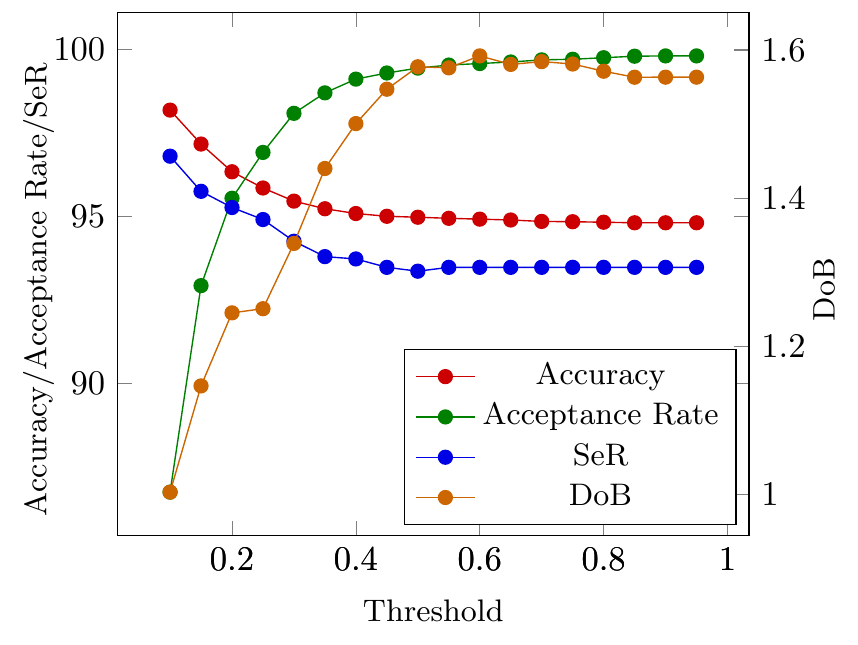}
    \caption{Model behaviour at various uncertainty thresholds}
    \label{fig:reject_option}
\end{figure}

Figure~\ref{fig:reject_option} shows the behavior of the model at different uncertainty thresholds. The choice of uncertainty threshold is application dependent. Although it is to be noted, at a threshold of $0.2$ the model rejects only $~4\%$ of all images and at the same time improving the overall accuracy $+2\%$ and reducing the DoB to $1.25$ from $1.62$.

\subsubsection{\textbf{Effectiveness on Other Biometric Modalities}} In order to analyze the generalizability of our method across different biometric modalities, we also conducted similar experiments on an ocular and a periocular dataset. For ocular analysis across gender, we used the VISOB~\cite{visob} dataset. Table ~\ref{tab:results_visob} shows the condensed results on VISOB dataset. The proposed method improved the accuracy as well as the DoB. For periocular analysis, we used the UFPR Periocular~\cite{ufpr} dataset. For this dataset, since race annotations are not available, we present the results across gender only. Table ~\ref{tab:results_ufpr} shows the results on UFPR dataset. As can be seen, the proposed method improved accuracy across gender even for ocular and periocular biometrics. 

\begin{table}[H]
    \centering
    \caption{Gender classification results of the proposed method across gender when trained and tested on VISOB dataset.}
    \label{tab:results_visob}
    \begin{tabular}{cccc} 
    \toprule
    \textbf{Config} & \textbf{Avg. Acc $\uparrow$} & \textbf{DoB $\downarrow$} & \textbf{SeR $\uparrow$} \\ \midrule
    Baseline &  87.95 & 15.27 & 48.37 \\
    NL & \textbf{89.17} & \textbf{14.18} & \textbf{52.51} \\
    \bottomrule
    \end{tabular}
\end{table}

\begin{table}[H]
    \centering
    \caption{Gender classification results of the proposed method across gender when trained and tested on UFPR dataset.}
    \label{tab:results_ufpr}
    \begin{tabular}{cccc} 
    \toprule
    \textbf{Config} & \textbf{Male $\uparrow$} & \textbf{Female $\uparrow$} & \textbf{Overall $\uparrow$} \\ \midrule
    Baseline & 96.13 & 95.13 & 95.60 \\
    NL & \textbf{97.13} & \textbf{95.88} & \textbf{96.47} \\
    \bottomrule
    \end{tabular}
\end{table}

\subsection{Ablation Study}
\subsubsection{Neighbour Size}
We observe that increasing the neighbor size of a sample improves the overall accuracy as well as the fairness of the model. For the ablation study, we conducted our experiments on the base NL configuration.

\begin{table}[H]
    \centering
    \caption{Ablation Study: Neighbour Size}
    \label{tab:ablate_neighbour}
    \begin{tabular}{cccc} 
    \toprule
    \textbf{Neighbour Size} & \textbf{Avg. Acc $\uparrow$} & \textbf{DoB $\downarrow$} & \textbf{SeR $\uparrow$} \\ \midrule
    1 & 94.35 & 1.95 & 92.20 \\
    3 & 94.58 & 1.69 & 92.69 \\
    5 & 94.50 & 1.73 & 93.39 \\
    7 & \textbf{94.67} & \textbf{1.67} & \textbf{93.78}\\
    \bottomrule
    \end{tabular}
    
\end{table}

\subsubsection{Lazy Regularization}
The lazy regularization value is chosen as a trade-off between training speed and the overall performance. The model works best when $n$ is $1$, i.e. no lazy regularization. Although this particular configuration is the best of all, we decided to stick with the $n=2$ configuration considering the speed trade-off.
\begin{table}[H]
    \centering
    \caption{Ablation Study : Lazy Regularization}
    \label{tab:ablate_lazy}
    \begin{tabular}{cccc} 
    \toprule
    \textbf{n} & \textbf{Avg. Acc $\uparrow$} & \textbf{DoB $\downarrow$} & \textbf{SeR $\uparrow$} \\ \midrule
    1 & \textbf{95.06} & \textbf{1.67} & 93.68 \\
    2 & 94.67 & \textbf{1.67} & \textbf{93.78} \\
    4 & 94.94 & 1.79 & 93.21 \\
    8 & 94.92 & 1.72 & 92.96 \\
    \bottomrule
    \end{tabular}
\end{table}

\subsubsection{Distance Metrics}
The distance function $\mathcal{L}_{\mathcal{N}}$ for neighbor regularization can be any function that can compute the distances between embedding. We evaluated L2 and Jensen-Shannon (JS) Divergence. Though cosine similarity and KL divergence are also possibilities. Experiments proved JS Divergence to be a better distance function for the model.
\begin{table}[H]
    \centering
    \caption{Ablation Study : Distance Metrics}
    \label{tab:ablate_dist}
    \begin{tabular}{cccc} 
    \toprule
    \textbf{Distance Metric} & \textbf{Avg. Acc $\uparrow$} & \textbf{DoB $\downarrow$} & \textbf{SeR $\uparrow$} \\ \midrule
    L2 & \textbf{94.76} & 1.71 & 92.91 \\
    JS Div & 94.67 & \textbf{1.67} & \textbf{93.78}\\
    \bottomrule
    \end{tabular}
\end{table}

\subsubsection{Effect of Generated views on Backpropagation}
The generated views are solely used for calculating the regularization term from the embeddings in the proposed approach, not for backpropagation or batch statistics calculation. We test this by using backpropagation to train a model using generated views, and we discover that the bias is substantially higher when generated views dominate the data distribution, despite the total accuracy improving marginally.
\begin{table}[H]
    \centering
    \caption{Ablation Study : With BackProp vs Without BackProp}
    \label{tab:ablate_backprop}
    \begin{tabular}{cccc} 
    \toprule
    \textbf{Config} & \textbf{Avg. Acc $\uparrow$} & \textbf{DoB $\downarrow$} & \textbf{SeR $\uparrow$} \\ \midrule
    With BackProp & \textbf{94.79} & 2.11 & 91.14 \\
    Without BackProp & 94.67 & \textbf{1.67} & \textbf{93.78}\\
    \bottomrule
    \end{tabular}
\end{table}

\section{Conclusion and Future Work}
Several studies have demonstrated that deep learning models can discriminate based on demographic attributes for biometric modalities such as face and ocular. Existing bias mitigation strategies often offer the trade-off between accuracy and fairness. In this study, we proposed a bias mitigation strategy that leverages the power of generative views and structured learning to learn invariant features that can improve the fairness and classification accuracy of a model simultaneously. We also propose a rejection mechanism based on uncertainty quantification that efficiently rejects uncertain samples at test time. Extensive evaluation across datasets and two biometric modalities demonstrate the generalizability of our proposed bias mitigation strategy to any biometric modality and classification task.  While the training process we currently use uses pregenerated neighbour samples, future research would explore ways to generate neighbor samples during the training phase to take into account the needs of the task at hand. Further, we will evaluate the efficacy of our proposed approach in mitigating bias for other computer vision tasks such as deepfake and biometric spoof attack detection.


\section{Acknowledgement}
This work is supported from National Science Foundation (NSF) award
no. 2129173. The research infrastructure used in this study is supported in part from a grant no. 13106715 from the Defense University Research Instrumentation Program (DURIP) from Air Force Office of Scientific Research.

\bibliographystyle{splncs04}
\bibliography{ref}

\begin{thebibliography}{10}
\providecommand{\url}[1]{\texttt{#1}}
\providecommand{\urlprefix}{URL }
\providecommand{\doi}[1]{https://doi.org/#1}

\bibitem{invert1}
Abdal, R., Qin, Y., Wonka, P.: Image2stylegan: How to embed images into the
  stylegan latent space? In: 2019 {IEEE/CVF} International Conference on
  Computer Vision, {ICCV} 2019, Seoul, Korea (South), October 27 - November 2,
  2019. pp. 4431--4440. {IEEE} (2019). \doi{10.1109/ICCV.2019.00453},
  \url{https://doi.org/10.1109/ICCV.2019.00453}

\bibitem{tradeoff1}
Adeli{-}Mosabbeb, E., Zhao, Q., Pfefferbaum, A., Sullivan, E.V., Fei{-}Fei, L.,
  Niebles, J.C., Pohl, K.M.: Representation learning with statistical
  independence to mitigate bias. In: {IEEE} Winter Conference on Applications
  of Computer Vision, {WACV} 2021, Waikoloa, HI, USA, January 3-8, 2021. pp.
  2512--2522. {IEEE} (2021). \doi{10.1109/WACV48630.2021.00256},
  \url{https://doi.org/10.1109/WACV48630.2021.00256}

\bibitem{bigan}
Brock, A., Donahue, J., Simonyan, K.: Large scale {GAN} training for high
  fidelity natural image synthesis. In: 7th International Conference on
  Learning Representations, {ICLR} 2019, New Orleans, LA, USA, May 6-9, 2019.
  OpenReview.net (2019), \url{https://openreview.net/forum?id=B1xsqj09Fm}

\bibitem{neural_graph_learning}
Bui, T.D., Ravi, S., Ramavajjala, V.: Neural graph learning: Training neural
  networks using graphs. In: Proceedings of the Eleventh ACM International
  Conference on Web Search and Data Mining. p. 64–71. WSDM '18, Association
  for Computing Machinery, New York, NY, USA (2018).
  \doi{10.1145/3159652.3159731}, \url{https://doi.org/10.1145/3159652.3159731}

\bibitem{gender_shades}
Buolamwini, J., Gebru, T.: Gender shades: Intersectional accuracy disparities
  in commercial gender classification. In: Friedler, S.A., Wilson, C. (eds.)
  Conference on Fairness, Accountability and Transparency, {FAT} 2018, 23-24
  February 2018, New York, NY, {USA}. Proceedings of Machine Learning Research,
  vol.~81, pp. 77--91. {PMLR} (2018),
  \url{http://proceedings.mlr.press/v81/buolamwini18a.html}

\bibitem{ensembling_views}
Chai, L., Zhu, J., Shechtman, E., Isola, P., Zhang, R.: Ensembling with deep
  generative views. In: {IEEE} Conference on Computer Vision and Pattern
  Recognition, {CVPR} 2021, virtual, June 19-25, 2021. pp. 14997--15007.
  Computer Vision Foundation / {IEEE} (2021),
  \url{https://openaccess.thecvf.com/content/CVPR2021/html/Chai_Ensembling_With_Deep_Generative_Views_CVPR_2021_paper.html}

\bibitem{fair_mixup}
Chuang, C., Mroueh, Y.: Fair mixup: Fairness via interpolation. In: 9th
  International Conference on Learning Representations, {ICLR} 2021, Virtual
  Event, Austria, May 3-7, 2021. OpenReview.net (2021),
  \url{https://openreview.net/forum?id=DNl5s5BXeBn}

\bibitem{editing1}
Collins, E., Bala, R., Price, B., S{\"{u}}sstrunk, S.: Editing in style:
  Uncovering the local semantics of gans. In: 2020 {IEEE/CVF} Conference on
  Computer Vision and Pattern Recognition, {CVPR} 2020, Seattle, WA, USA, June
  13-19, 2020. pp. 5770--5779. Computer Vision Foundation / {IEEE} (2020).
  \doi{10.1109/CVPR42600.2020.00581}

\bibitem{invert2}
Creswell, A., Bharath, A.A.: Inverting the generator of a generative
  adversarial network. {IEEE} Trans. Neural Networks Learn. Syst.
  \textbf{30}(7),  1967--1974 (2019). \doi{10.1109/TNNLS.2018.2875194},
  \url{https://doi.org/10.1109/TNNLS.2018.2875194}

\bibitem{multitask}
Das, A., Dantcheva, A., Br{\'{e}}mond, F.: Mitigating bias in gender, age and
  ethnicity classification: {A} multi-task convolution neural network approach.
  In: Leal{-}Taix{\'{e}}, L., Roth, S. (eds.) Computer Vision - {ECCV} 2018
  Workshops - Munich, Germany, September 8-14, 2018, Proceedings, Part {I}.
  Lecture Notes in Computer Science, vol. 11129, pp. 573--585. Springer (2018).
  \doi{10.1007/978-3-030-11009-3\_35},
  \url{https://doi.org/10.1007/978-3-030-11009-3_35}

\bibitem{walking_bias}
Denton, E., Hutchinson, B., Mitchell, M., Gebru, T.: Detecting bias with
  generative counterfactual face attribute augmentation. CoRR
  \textbf{abs/1906.06439} (2019), \url{http://arxiv.org/abs/1906.06439}

\bibitem{ganalyze}
Goetschalckx, L., Andonian, A., Oliva, A., Isola, P.: Ganalyze: Toward visual
  definitions of cognitive image properties. In: 2019 {IEEE/CVF} International
  Conference on Computer Vision, {ICCV} 2019, Seoul, Korea (South), October 27
  - November 2, 2019. pp. 5743--5752. {IEEE} (2019).
  \doi{10.1109/ICCV.2019.00584}, \url{https://doi.org/10.1109/ICCV.2019.00584}

\bibitem{dob}
Gong, S., Liu, X., Jain, A.K.: Debface: De-biasing face recognition. CoRR
  \textbf{abs/1911.08080} (2019), \url{http://arxiv.org/abs/1911.08080}

\bibitem{gan}
Goodfellow, I., Pouget-Abadie, J., Mirza, M., Xu, B., Warde-Farley, D., Ozair,
  S., Courville, A., Bengio, Y.: Generative adversarial nets. In: Ghahramani,
  Z., Welling, M., Cortes, C., Lawrence, N., Weinberger, K.Q. (eds.) Advances
  in Neural Information Processing Systems. vol.~27. Curran Associates, Inc.
  (2014),
  \url{https://proceedings.neurips.cc/paper/2014/file/5ca3e9b122f61f8f06494c97b1afccf3-Paper.pdf}

\bibitem{goodfellow_adverarial}
Goodfellow, I.J., Shlens, J., Szegedy, C.: Explaining and harnessing
  adversarial examples. In: Bengio, Y., LeCun, Y. (eds.) 3rd International
  Conference on Learning Representations, {ICLR} 2015, San Diego, CA, USA, May
  7-9, 2015, Conference Track Proceedings (2015),
  \url{http://arxiv.org/abs/1412.6572}

\bibitem{invert4}
Guan, S., Tai, Y., Ni, B., Zhu, F., Huang, F., Yang, X.: Collaborative learning
  for faster stylegan embedding. CoRR  \textbf{abs/2007.01758} (2020),
  \url{https://arxiv.org/abs/2007.01758}

\bibitem{editing2}
H{\"{a}}rk{\"{o}}nen, E., Hertzmann, A., Lehtinen, J., Paris, S.: Ganspace:
  Discovering interpretable {GAN} controls. In: Larochelle, H., Ranzato, M.,
  Hadsell, R., Balcan, M., Lin, H. (eds.) Advances in Neural Information
  Processing Systems 33: Annual Conference on Neural Information Processing
  Systems 2020, NeurIPS 2020, December 6-12, 2020, virtual (2020),
  \url{https://proceedings.neurips.cc/paper/2020/hash/6fe43269967adbb64ec6149852b5cc3e-Abstract.html}

\bibitem{fid}
Heusel, M., Ramsauer, H., Unterthiner, T., Nessler, B., Hochreiter, S.: Gans
  trained by a two time-scale update rule converge to a local nash equilibrium.
  In: Guyon, I., Luxburg, U.V., Bengio, S., Wallach, H., Fergus, R.,
  Vishwanathan, S., Garnett, R. (eds.) Advances in Neural Information
  Processing Systems. vol.~30. Curran Associates, Inc. (2017),
  \url{https://proceedings.neurips.cc/paper/2017/file/8a1d694707eb0fefe65871369074926d-Paper.pdf}

\bibitem{DBLP:conf/fgr/JainH04}
Jain, A., Huang, J.: Integrating independent components and linear discriminant
  analysis for gender classification. In: Sixth {IEEE} International Conference
  on Automatic Face and Gesture Recognition {(FGR} 2004), May 17-19, 2004,
  Seoul, Korea. pp. 159--163. {IEEE} Computer Society (2004).
  \doi{10.1109/AFGR.2004.1301524},
  \url{https://doi.org/10.1109/AFGR.2004.1301524}

\bibitem{reject_option}
Kamiran, F., Karim, A., Zhang, X.: Decision theory for discrimination-aware
  classification. In: 2012 IEEE 12th International Conference on Data Mining.
  pp. 924--929 (2012). \doi{10.1109/ICDM.2012.45}

\bibitem{progan}
Karras, T., Aila, T., Laine, S., Lehtinen, J.: Progressive growing of gans for
  improved quality, stability, and variation. In: 6th International Conference
  on Learning Representations, {ICLR} 2018, Vancouver, BC, Canada, April 30 -
  May 3, 2018, Conference Track Proceedings. OpenReview.net (2018),
  \url{https://openreview.net/forum?id=Hk99zCeAb}

\bibitem{stylegan2-ada}
Karras, T., Aittala, M., Hellsten, J., Laine, S., Lehtinen, J., Aila, T.:
  Training generative adversarial networks with limited data. In: Larochelle,
  H., Ranzato, M., Hadsell, R., Balcan, M., Lin, H. (eds.) Advances in Neural
  Information Processing Systems 33: Annual Conference on Neural Information
  Processing Systems 2020, NeurIPS 2020, December 6-12, 2020, virtual (2020),
  \url{https://proceedings.neurips.cc/paper/2020/hash/8d30aa96e72440759f74bd2306c1fa3d-Abstract.html}

\bibitem{stylegan}
Karras, T., Laine, S., Aila, T.: A style-based generator architecture for
  generative adversarial networks. In: {IEEE} Conference on Computer Vision and
  Pattern Recognition, {CVPR} 2019, Long Beach, CA, USA, June 16-20, 2019. pp.
  4401--4410. Computer Vision Foundation / {IEEE} (2019).
  \doi{10.1109/CVPR.2019.00453}

\bibitem{stylegan2}
Karras, T., Laine, S., Aittala, M., Hellsten, J., Lehtinen, J., Aila, T.:
  Analyzing and improving the image quality of stylegan. In: 2020 {IEEE/CVF}
  Conference on Computer Vision and Pattern Recognition, {CVPR} 2020, Seattle,
  WA, USA, June 13-19, 2020. pp. 8107--8116. Computer Vision Foundation /
  {IEEE} (2020). \doi{10.1109/CVPR42600.2020.00813}

\bibitem{keyes}
Keyes, O.: The misgendering machines: Trans/hci implications of automatic
  gender recognition. Proc. {ACM} Hum. Comput. Interact.  \textbf{2}({CSCW}),
  88:1--88:22 (2018). \doi{10.1145/3274357},
  \url{https://doi.org/10.1145/3274357}

\bibitem{article1}
Khan, S., Ahmad, M., Nazir, M., Riaz, N.: A comparative analysis of gender
  classification techniques. Middle - East Journal of Scientific Research
  \textbf{20},  1--13 (01 2014). \doi{10.5829/idosi.mejsr.2014.20.01.11434}

\bibitem{ssl_gcn}
Kipf, T.N., Welling, M.: Semi-supervised classification with graph
  convolutional networks. In: 5th International Conference on Learning
  Representations, {ICLR} 2017, Toulon, France, April 24-26, 2017, Conference
  Track Proceedings. OpenReview.net (2017),
  \url{https://openreview.net/forum?id=SJU4ayYgl}

\bibitem{bias_fairface}
Krishnan, A., Almadan, A., Rattani, A.: Understanding fairness of gender
  classification algorithms across gender-race groups. In: Wani, M.A., Luo, F.,
  Li, X.A., Dou, D., Bonchi, F. (eds.) 19th {IEEE} International Conference on
  Machine Learning and Applications, {ICMLA} 2020, Miami, FL, USA, December
  14-17, 2020. pp. 1028--1035. {IEEE} (2020).
  \doi{10.1109/ICMLA51294.2020.00167},
  \url{https://doi.org/10.1109/ICMLA51294.2020.00167}

\bibitem{9717383}
Krishnan, A., Almadan, A., Rattani, A.: Investigating fairness of ocular
  biometrics among young, middle-aged, and older adults. In: 2021 International
  Carnahan Conference on Security Technology (ICCST). pp.~1--7 (2021).
  \doi{10.1109/ICCST49569.2021.9717383}

\bibitem{10.1007/978-3-030-68793-9_16}
Krishnan, A., Almadan, A., Rattani, A.: Probing fairness of mobile ocular
  biometrics methods across gender on visob 2.0 dataset. In: Del~Bimbo, A.,
  Cucchiara, R., Sclaroff, S., Farinella, G.M., Mei, T., Bertini, M.,
  Escalante, H.J., Vezzani, R. (eds.) Pattern Recognition. ICPR International
  Workshops and Challenges. pp. 229--243. Springer International Publishing,
  Cham (2021)

\bibitem{fairface}
Kärkkäinen, K., Joo, J.: Fairface: Face attribute dataset for balanced race,
  gender, and age (2019)

\bibitem{ser}
Lin, F., Wu, Y., Zhuang, Y., Long, X., Xu, W.: Human gender classification: a
  review. Int. J. Biom.  \textbf{8}(3/4),  275--300 (2016).
  \doi{10.1504/IJBM.2016.10003589},
  \url{https://doi.org/10.1504/IJBM.2016.10003589}

\bibitem{invert3}
Lipton, Z.C., Tripathi, S.: Precise recovery of latent vectors from generative
  adversarial networks. In: 5th International Conference on Learning
  Representations, {ICLR} 2017, Toulon, France, April 24-26, 2017, Workshop
  Track Proceedings. OpenReview.net (2017),
  \url{https://openreview.net/forum?id=HJC88BzFl}

\bibitem{lp_insta}
Mahajan, D., Girshick, R.B., Ramanathan, V., He, K., Paluri, M., Li, Y.,
  Bharambe, A., van~der Maaten, L.: Exploring the limits of weakly supervised
  pretraining. In: Ferrari, V., Hebert, M., Sminchisescu, C., Weiss, Y. (eds.)
  Computer Vision - {ECCV} 2018 - 15th European Conference, Munich, Germany,
  September 8-14, 2018, Proceedings, Part {II}. Lecture Notes in Computer
  Science, vol. 11206, pp. 185--201. Springer (2018).
  \doi{10.1007/978-3-030-01216-8\_12},
  \url{https://doi.org/10.1007/978-3-030-01216-8_12}

\bibitem{att_aware_debias}
Majumdar, P., Singh, R., Vatsa, M.: Attention aware debiasing for unbiased
  model prediction. In: {IEEE/CVF} International Conference on Computer Vision
  Workshops, {ICCVW} 2021, Montreal, BC, Canada, October 11-17, 2021. pp.
  4116--4124. {IEEE} (2021). \doi{10.1109/ICCVW54120.2021.00459},
  \url{https://doi.org/10.1109/ICCVW54120.2021.00459}

\bibitem{miyato}
Miyato, T., Maeda, S.i., Koyama, M., Ishii, S.: Virtual adversarial training: A
  regularization method for supervised and semi-supervised learning. IEEE
  Transactions on Pattern Analysis and Machine Intelligence  \textbf{PP} (04
  2017). \doi{10.1109/TPAMI.2018.2858821}

\bibitem{diveface}
Morales, A., Fi{\'{e}}rrez, J., Vera{-}Rodr{\'{\i}}guez, R., Tolosana, R.:
  Sensitivenets: Learning agnostic representations with application to face
  images. {IEEE} Trans. Pattern Anal. Mach. Intell.  \textbf{43}(6),
  2158--2164 (2021). \doi{10.1109/TPAMI.2020.3015420},
  \url{https://doi.org/10.1109/TPAMI.2020.3015420}

\bibitem{bias_color}
Muthukumar, V.: Color-theoretic experiments to understand unequal gender
  classification accuracy from face images. In: {IEEE} Conference on Computer
  Vision and Pattern Recognition Workshops, {CVPR} Workshops 2019, Long Beach,
  CA, USA, June 16-20, 2019. pp. 2286--2295. Computer Vision Foundation /
  {IEEE} (2019). \doi{10.1109/CVPRW.2019.00282}

\bibitem{BiasDeep}
Nadimpalli, A.V., Rattani, A.: {GBDF}: Gender balanced deepfake dataset towards
  fair deepfake detection (2022). \doi{10.48550/ARXIV.2207.10246},
  \url{https://arxiv.org/abs/2207.10246}

\bibitem{att_aware_filter}
Nagpal, S., Singh, M., Singh, R., Vatsa, M.: Attribute aware filter-drop for
  bias-invariant classification. In: 2020 IEEE/CVF Conference on Computer
  Vision and Pattern Recognition Workshops (CVPRW). pp. 147--153 (2020).
  \doi{10.1109/CVPRW50498.2020.00024}

\bibitem{nist_gender}
Ngan, M., Grother, P.: Face recognition vendor test (frvt) - performance of
  automated gender classification algorithms (2015-04-20 2015).
  \doi{https://doi.org/10.6028/NIST.IR.8052}

\bibitem{visob}
Nguyen, H.M., Reddy, N., Rattani, A., Derakhshani, R.: {VISOB} 2.0 - the second
  international competition on mobile ocular biometric recognition. In: Bimbo,
  A.D., Cucchiara, R., Sclaroff, S., Farinella, G.M., Mei, T., Bertini, M.,
  Escalante, H.J., Vezzani, R. (eds.) Pattern Recognition. {ICPR} International
  Workshops and Challenges - Virtual Event, January 10-15, 2021, Proceedings,
  Part {VIII}. Lecture Notes in Computer Science, vol. 12668, pp. 200--208.
  Springer (2020). \doi{10.1007/978-3-030-68793-9\_14},
  \url{https://doi.org/10.1007/978-3-030-68793-9\_14}

\bibitem{readme}
Park, S., Kim, D., Hwang, S., Byun, H.: {README:} representation learning by
  fairness-aware disentangling method. CoRR  \textbf{abs/2007.03775} (2020),
  \url{https://arxiv.org/abs/2007.03775}

\bibitem{invert5}
Perarnau, G., van~de Weijer, J., Raducanu, B.C., {\'{A}}lvarez, J.M.:
  Invertible conditional gans for image editing. CoRR  \textbf{abs/1611.06355}
  (2016), \url{http://arxiv.org/abs/1611.06355}

\bibitem{clip}
Radford, A., Kim, J.W., Hallacy, C., Ramesh, A., Goh, G., Agarwal, S., Sastry,
  G., Askell, A., Mishkin, P., Clark, J., Krueger, G., Sutskever, I.: Learning
  transferable visual models from natural language supervision. CoRR
  \textbf{abs/2103.00020} (2021), \url{https://arxiv.org/abs/2103.00020}

\bibitem{gan1}
Radford, A., Metz, L., Chintala, S.: Unsupervised representation learning with
  deep convolutional generative adversarial networks. In: Bengio, Y., LeCun, Y.
  (eds.) 4th International Conference on Learning Representations, {ICLR} 2016,
  San Juan, Puerto Rico, May 2-4, 2016, Conference Track Proceedings (2016),
  \url{http://arxiv.org/abs/1511.06434}

\bibitem{commercial_bias}
Raji, I.D., Buolamwini, J.: Actionable auditing: Investigating the impact of
  publicly naming biased performance results of commercial {AI} products. In:
  Conitzer, V., Hadfield, G.K., Vallor, S. (eds.) Proceedings of the 2019
  {AAAI/ACM} Conference on AI, Ethics, and Society, {AIES} 2019, Honolulu, HI,
  USA, January 27-28, 2019. pp. 429--435. {ACM} (2019).
  \doi{10.1145/3306618.3314244}, \url{https://doi.org/10.1145/3306618.3314244}

\bibitem{gan_debias}
Ramaswamy, V.V., Kim, S.S.Y., Russakovsky, O.: Fair attribute classification
  through latent space de-biasing. In: {IEEE} Conference on Computer Vision and
  Pattern Recognition, {CVPR} 2021, virtual, June 19-25, 2021. pp. 9301--9310.
  Computer Vision Foundation / {IEEE} (2021),
  \url{https://openaccess.thecvf.com/content/CVPR2021/html/Ramaswamy_Fair_Attribute_Classification_Through_Latent_Space_De-Biasing_CVPR_2021_paper.html}

\bibitem{bowker}
Randall, D.W.: Geoffrey bowker and susan leigh star, sorting things out:
  Classification and its consequences - review. Comput. Support. Cooperative
  Work.  \textbf{10}(1),  147--153 (2001). \doi{10.1023/A:1011229919958},
  \url{https://doi.org/10.1023/A:1011229919958}

\bibitem{morph}
Ricanek, K., Tesafaye, T.: Morph: a longitudinal image database of normal adult
  age-progression. In: 7th International Conference on Automatic Face and
  Gesture Recognition (FGR06). pp. 341--345 (2006). \doi{10.1109/FGR.2006.78}

\bibitem{psp}
Richardson, E., Alaluf, Y., Patashnik, O., Nitzan, Y., Azar, Y., Shapiro, S.,
  Cohen{-}Or, D.: Encoding in style: {A} stylegan encoder for image-to-image
  translation. In: {IEEE} Conference on Computer Vision and Pattern
  Recognition, {CVPR} 2021, virtual, June 19-25, 2021. pp. 2287--2296. Computer
  Vision Foundation / {IEEE} (2021),
  \url{https://openaccess.thecvf.com/content/CVPR2021/html/Richardson_Encoding_in_Style_A_StyleGAN_Encoder_for_Image-to-Image_Translation_CVPR_2021_paper.html}

\bibitem{Ryu18}
Ryu, H.J., Adam, H., Mitchell, M.: Inclusivefacenet: Improving face attribute
  detection with race and gender diversity. In: Workshop on Fairness,
  Accountability, and Transparency in Machine Learning. pp.~1--6 (March 2018)

\bibitem{edl}
Sensoy, M., Kaplan, L.M., Kandemir, M.: Evidential deep learning to quantify
  classification uncertainty. In: Bengio, S., Wallach, H.M., Larochelle, H.,
  Grauman, K., Cesa{-}Bianchi, N., Garnett, R. (eds.) Advances in Neural
  Information Processing Systems 31: Annual Conference on Neural Information
  Processing Systems 2018, NeurIPS 2018, December 3-8, 2018, Montr{\'{e}}al,
  Canada. pp. 3183--3193 (2018),
  \url{https://proceedings.neurips.cc/paper/2018/hash/a981f2b708044d6fb4a71a1463242520-Abstract.html}

\bibitem{walking1}
Shen, Y., Gu, J., Tang, X., Zhou, B.: Interpreting the latent space of gans for
  semantic face editing. In: 2020 {IEEE/CVF} Conference on Computer Vision and
  Pattern Recognition, {CVPR} 2020, Seattle, WA, USA, June 13-19, 2020. pp.
  9240--9249. Computer Vision Foundation / {IEEE} (2020).
  \doi{10.1109/CVPR42600.2020.00926}

\bibitem{sefa}
Shen, Y., Zhou, B.: Closed-form factorization of latent semantics in gans. In:
  {IEEE} Conference on Computer Vision and Pattern Recognition, {CVPR} 2021,
  virtual, June 19-25, 2021. pp. 1532--1540. Computer Vision Foundation /
  {IEEE} (2021),
  \url{https://openaccess.thecvf.com/content/CVPR2021/html/Shen_Closed-Form_Factorization_of_Latent_Semantics_in_GANs_CVPR_2021_paper.html}

\bibitem{Siddiqui_2022_CVPR}
Siddiqui, H., Rattani, A., Ricanek, K., Hill, T.: An examination of bias of
  facial analysis based bmi prediction models. In: Proceedings of the IEEE/CVF
  Conference on Computer Vision and Pattern Recognition (CVPR) Workshops. pp.
  2926--2935 (June 2022)

\bibitem{chlng}
Singh, R., Majumdar, P., Mittal, S., Vatsa, M.: Anatomizing bias in facial
  analysis. CoRR  \textbf{abs/2112.06522} (2021),
  \url{https://arxiv.org/abs/2112.06522}

\bibitem{gender_age_transfer}
Smith, P., Chen, C.: Transfer learning with deep cnns for gender recognition
  and age estimation (2018)

\bibitem{tartaglione}
Tartaglione, E., Barbano, C.A., Grangetto, M.: End: Entangling and
  disentangling deep representations for bias correction. In: {IEEE} Conference
  on Computer Vision and Pattern Recognition, {CVPR} 2021, virtual, June 19-25,
  2021. pp. 13508--13517. Computer Vision Foundation / {IEEE} (2021)

\bibitem{nsl}
Tensorflow: The neural structured learning framework,
  \url{https://www.tensorflow.org/neural_structured_learning/framework}

\bibitem{e4e}
Tov, O., Alaluf, Y., Nitzan, Y., Patashnik, O., Cohen{-}Or, D.: Designing an
  encoder for stylegan image manipulation. {ACM} Trans. Graph.  \textbf{40}(4),
   133:1--133:14 (2021). \doi{10.1145/3450626.3459838},
  \url{https://doi.org/10.1145/3450626.3459838}

\bibitem{10.1145/3194770.3194776}
Verma, S., Rubin, J.: Fairness definitions explained. In: Proceedings of the
  International Workshop on Software Fairness. p. 1–7. FairWare '18,
  Association for Computing Machinery, New York, NY, USA (2018).
  \doi{10.1145/3194770.3194776}, \url{https://doi.org/10.1145/3194770.3194776}

\bibitem{walking_un1}
Voynov, A., Babenko, A.: Unsupervised discovery of interpretable directions in
  the {GAN} latent space. In: Proceedings of the 37th International Conference
  on Machine Learning, {ICML} 2020, 13-18 July 2020, Virtual Event. Proceedings
  of Machine Learning Research, vol.~119, pp. 9786--9796. {PMLR} (2020),
  \url{http://proceedings.mlr.press/v119/voynov20a.html}

\bibitem{walking_un2}
Wang, B., Ponce, C.R.: A geometric analysis of deep generative image models and
  its applications. In: 9th International Conference on Learning
  Representations, {ICLR} 2021, Virtual Event, Austria, May 3-7, 2021.
  OpenReview.net (2021), \url{https://openreview.net/forum?id=GH7QRzUDdXG}

\bibitem{tradeoff2}
Wang, T., Zhao, J., Yatskar, M., Chang, K., Ordonez, V.: Balanced datasets are
  not enough: Estimating and mitigating gender bias in deep image
  representations. In: 2019 {IEEE/CVF} International Conference on Computer
  Vision, {ICCV} 2019, Seoul, Korea (South), October 27 - November 2, 2019. pp.
  5309--5318. {IEEE} (2019). \doi{10.1109/ICCV.2019.00541},
  \url{https://doi.org/10.1109/ICCV.2019.00541}

\bibitem{wayman1997large}
Wayman, J.: Large-scale civilian biometric systems-issues and feasibility. In:
  Proceedings of Card Tech/Secur Tech ID. vol.~732 (1997)

\bibitem{jsdiv}
Wikipedia: Jensen–shannon divergence (May 2022),
  \url{https://en.wikipedia.org/wiki/Jensen-Shannon_divergence}

\bibitem{editing4}
Wu, Z., Lischinski, D., Shechtman, E.: Stylespace analysis: Disentangled
  controls for stylegan image generation. In: {IEEE} Conference on Computer
  Vision and Pattern Recognition, {CVPR} 2021, virtual, June 19-25, 2021. pp.
  12863--12872. Computer Vision Foundation / {IEEE} (2021)

\bibitem{ufpr}
Zanlorensi, L.A., Laroca, R., Lucio, D.R., Santos, L.R., Britto, A.S., Menotti,
  D.: Ufpr-periocular: A periocular dataset collected by mobile devices in
  unconstrained scenarios (2020). \doi{10.48550/ARXIV.2011.12427},
  \url{https://arxiv.org/abs/2011.12427}

\bibitem{mtcnn}
Zhang, K., Zhang, Z., Li, Z., Qiao, Y.: Joint face detection and alignment
  using multi-task cascaded convolutional networks. CoRR
  \textbf{abs/1604.02878} (2016), \url{http://arxiv.org/abs/1604.02878}

\bibitem{utkface}
Zhang, Z., Song, Y., , Qi, H.: Age progression/regression by conditional
  adversarial autoencoder. In: IEEE Conference on Computer Vision and Pattern
  Recognition (CVPR). IEEE (2017)

\end{thebibliography}
\end{document}